# Open Set Recognition for Endoscopic Image Classification: A Deep Learning Approach on the Kvasir Dataset

**Seoyoun Son**[†]
Emory University
rachel4175@gmail.com

**John Lin**[†]
Emory University
john.lin2@emory.edu

**Sun Min Lee**[†]
Boston University
leeclara@bu.edu

**Daniel Son**[†]
Gretchen Whitney High School
daniel_son@myabcusd.org

**Hayeon Lee**[†]
Auburn High School
haylee@acsk12.org

**Jeongho Lee**
Hanyang University ERICA
ghdi1ghdi@hanyang.ac.kr

**Seongji Lee**
Hanbat National University
glasrooml4566@gmail.com

**Kasra Moazzami**[*]
Emory University
kasra.moazzami@emory.edu

† These authors contributed equally to this work.

* Corresponding author: kasra.moazzami@emory.edu

## Abstract

Endoscopic image classification plays a pivotal role in medical diagnostics by identifying anatomical landmarks and pathological findings. However, conventional closed-set classification frameworks are inherently limited in open-world clinical settings, where previously unseen conditions can arise andcompromise model reliability. To address this, we explore the application of Open Set Recognition (OSR) techniques on the Kvasir dataset, a publicly available and diverse endoscopic image collection. In this study, we evaluate and compare the OSR capabilities of several representative deep learning architectures, including ResNet-50, Swin Transformer, and a hybrid ResNet-Transformer model, under both closed-set and open-set conditions. OpenMax is adopted as a baseline OSR method to assess the ability of these models to distinguish known classes from previously unseen categories. This work represents one of the first efforts to apply open set recognition to the Kvasir dataset and provides a foundational benchmark for evaluating OSR performance in medical image analysis. Our results offer practical insights into model behavior in clinically realistic settings and highlight the importance of OSR techniques for the safe deployment of AI systems in endoscopy.

## 1. Introduction

Recent advancements in deep learning have substantially improved the performance of medical image analysis, enabling more accurate, efficient, and automated diagnostic processes [1]. Particularly, convolutional neural networks (CNNs), a class of deep learning models, have demonstrated outstanding capabilities in analyzing medical imaging modalities such as endoscopy, radiography, magnetic resonance imaging (MRI), computed tomography (CT), and ultrasound [2,3] These developments hold significant clinical implications, including early detection and accurate classification of diseases, assistance in treatment decision-making, and reduction of diagnostic errors caused by human variability or fatigue [2]. As a result, integrating deep learning-based medical image analysis into clinical workflows has been recognized as a critical step toward improving healthcare quality and patient outcomes [3].

Endoscopic image analysis plays a pivotal for treating gastrointestinal (GI) diseases as well; significantly improving clinical outcomes and patient survival rates [4]. However, analyzing GI endoscopic images presents considerable challenges due to inherent heterogeneity in appearance caused by varying anatomical structures, diverse pathological conditions, and inconsistent illumination or imaging protocols [5]. Furthermore, different gastrointestinal conditions frequently exhibit visual similarity, making accurate discrimination between normal and abnormal findings, as well as between closely related pathological categories, exceptionally challenging [6]. These challenges exemplify the need for robust and adaptable image analysis techniques capable of addressing the complexities inherent in GI endoscopic datasets such as the Kvasir dataset [5].

Despite the promising advances achieved by recent deep learning models, conventional closed-set classification paradigms remain fundamentally constrained in their applicability to real-world medical imaging tasks. Prior works em- ploying CNN-based architectures have demonstrated considerable success in recognizing known anatomical structures and pathological patterns within well-curated, labeled datasets [7]. Additionally, the introduction of Transformer-based models has offered enhanced global context modeling and improved representation of complex spatial dependencies [8]. However, these models exhibit a critical vulnerability when confronted with out-of-distribution samples or new classes that were absent during training—a situation that is not only common but inevitable in clinical practice [9].

This challenge is not unique to medical imaging. In the field of surveillance, for instance, it has been demonstrated that anomaly detection systems must be able to recognize and reject previously unseen events that deviate from learned normal behavior [10]. This study on video anomaly identification highlights the importance of open-set recognition in real-world scenarios where encountering unknown or out-of-distribution data is inevitable.

Traditional classifiers are architecturally predisposed to assign one of the known class labels to any input, regardless of its semantic alignment, often resulting in overconfident misclassifications when encountering unfamiliar cases [11]. This limitation becomes particularly pronounced in gastrointestinal endoscopy, where fine-grained distinctions between morphologically similar classes—such as early-stage inflammation versus normal mucosa—pose a significant challenge [12]. Moreover, datasets used in training are often biased toward prevalent or well-defined pathologies, thereby neglecting rare, emerging, or ambiguous cases [13]. Such imbalance impairs the model's capacity to generalize and undermines diagnostic reliability, especially in safety-critical scenarios where erroneous predictions may carry severe clinical consequences [14].

In practical clinical scenarios, it's often challenging or even impossible to acquire labeled training samples that encompass all possible pathological findings or anatomical structures encountered during medical diagnosis. Specifically, medical datasets inherently contain unknown or unseen classes due to the continuous emergence of novel diseases, diverse variations in anatomical structures, and differences in image capturing devices or conditions. Consequently, conventional classification approaches, which typically assume a closed-set scenario—that is, training and testing sets sharing identical class distributions—may fail or provide misleading results when presented with classes not observed during training [1]. To address this issue, Open Set Recognition (OSR) has emerged as a crucial area of research, aiming to accurately identify samples belonging to unknown classes, thereby enhancing the robustness and reliability of diagnostic systems deployed in real-world medical applications [9].

## 2. Kvasir dataset

The Kvasir dataset [1] comprises high-quality images of the gastrointestinal (GI) tract, curated specifically for multi- class classification tasks. It contains a total of 8,000 images, evenly distributed across eight categories, with 1,000 images per class. These categories include anatomical landmarks (pylorus, Z-line, and cecum), pathological conditions (esophagitis, ulcerative colitis, and polyps), and procedural findings (dyed-lifted polyps and dyed resection margins). This well-structured composition ensures balanced representation of both normal and abnormal GI conditions, making the dataset a robust and reliable benchmark for training and evaluating deep learning models in gastrointestinal image classification.

## 3. Methodology

**3.1 Swin Transformer**

The Swin Transformer [15] is adopted as the feature extraction backbone for its ability to efficiently capture both local and global contextual information. As a hierarchical vision transformer, it performs self-attention within non- overlapping local windows and introduces cross-window connections via a shifted window mechanism. This architecture enables effective multi-scale representation learning, which is particularly advantageous for high-resolution medical images such as endoscopic data, where both fine-grained details and broader contextual understanding are critical. The hierarchical design starts with small image patches and progressively merges them, forming a feature pyramid well-suited for dense prediction tasks like segmentation and classification. The use of shifted windows at each stage allows the model to capture inter-window dependencies, thereby addressing the locality limitations of traditional window-based attention. In our approach, the multi-level features extracted by the Swin Transformer are leveraged to improve the performance of the open set recognition (OSR) module by enhancing the model's capacity to distinguish between known and unknown gastrointestinal conditions.

### 3.2 ResNet

ResNet[16] is utilized as a baseline convolutional neural network (CNN) model to compare the effectiveness of transformer-based architectures. ResNet's residual learning mechanism enables efficient training of deep networks by mitigating the vanishing gradient problem through identity-based skip connections. These residual blocks allow for stable gradient propagation and support hierarchical feature extraction, which are essential in processing high-resolution medical images. We employ the ResNet-50 variant, which consists of bottleneck residual blocks structured into a four-stage hierarchy. The architecture serves as a comparative baseline for evaluating feature representations in our open set recognition (OSR) framework. Its proven robustness across various medical image analysis tasks, including classification of gastrointestinal conditions, provides a strong foundation for benchmarking transformer-based approaches such as Swin Transformer.

### 3.3 HybridSwinResNet for Open Set Recognition

To address the limitations of using either CNNs or Transformers in isolation, we propose a hybrid architecture, Hy- bridSwinResNet, tailored for open set recognition (OSR) in endoscopic image classification. This model integrates a Swin Transformer as the primary feature extractor and incorporates ResNet-inspired residual blocks as a refinement mod- ule to enhance local feature representation. The Swin Transformer component captures hierarchical global contextual information via shifted-window self-attention, while the residual refinement path—adapted from ResNet—reinforces local texture and pattern recognition. By combining these heterogeneous feature representations, the model forms a unified multi-scale embedding space that supports both fine-grained detail and semantic structure modeling. The fused features are projected through a fully connected layer to map high-dimensional representations into a compact latent space optimized for class separability. The final open set recognition process is implemented using a joint loss function that combines supervised classification for known classes with an anomaly detection mechanism for identifying unseen instances. This hybrid architecture enhances discriminative capacity and generalization by leveraging both localized filters and global attention, making it especially robust in high-variability settings like gastrointestinal endoscopy, where unknown pathological patterns are frequently encountered.

Similar to our hybrid Swin-ResNet design, prior research has shown that incorporating diverse architectural elements or external contextual information can significantly improve a model's generalization to unseen inputs. For example, a fully- convolutional network tailored for background subtraction in completely unseen video scenes demonstrated robust open- set performance by leveraging semantic context and data augmentation [17]. Moreover, hybrid generative-discriminative approaches—combining a variational autoencoder with a GAN to disentangle expression from identity—have proven effective in open-set scenarios. By learning a compact, invariant feature space, these methods enable reliable recognition on previously unseen individuals [18], an insight that directly parallels our motivation for uniting CNN-based local features with transformer-based global representations.

## 4. Experimental setup

### 4.1 Dataset Preparation

Experiments were conducted using the Kvasir dataset, which consists of 8,000 RGB endoscopic images uniformly distributed across eight categories: dyed-lifted-polyps, dyed-resection-margins, esophagitis, normal-cecum, normal- pylorus, normal-z-line, polyps, and ulcerative-colitis. To ensure input consistency and compatibility with standard deep learning architectures, all images were resized to $224 \times 224$ pixels. The dataset was partitioned into training (70%), validation (10%), and test (20%) subsets using stratified sampling, thereby maintaining the original class distribution across each split. To enhance model generalization and account for the variability inherent in clinical endoscopic imaging, a data augmentation pipeline was applied to the training set. This included random horizontal flipping, rotations up to 10 degrees, and color jittering to simulate changes in brightness and contrast. These augmentations help the model learn invariance to spatial and illumination-related perturbations. After augmentation, all images were normalized using the ImageNet standard mean and standard deviation values. The preprocessing and augmentation steps were applied consistently during training to every image in the training subset.

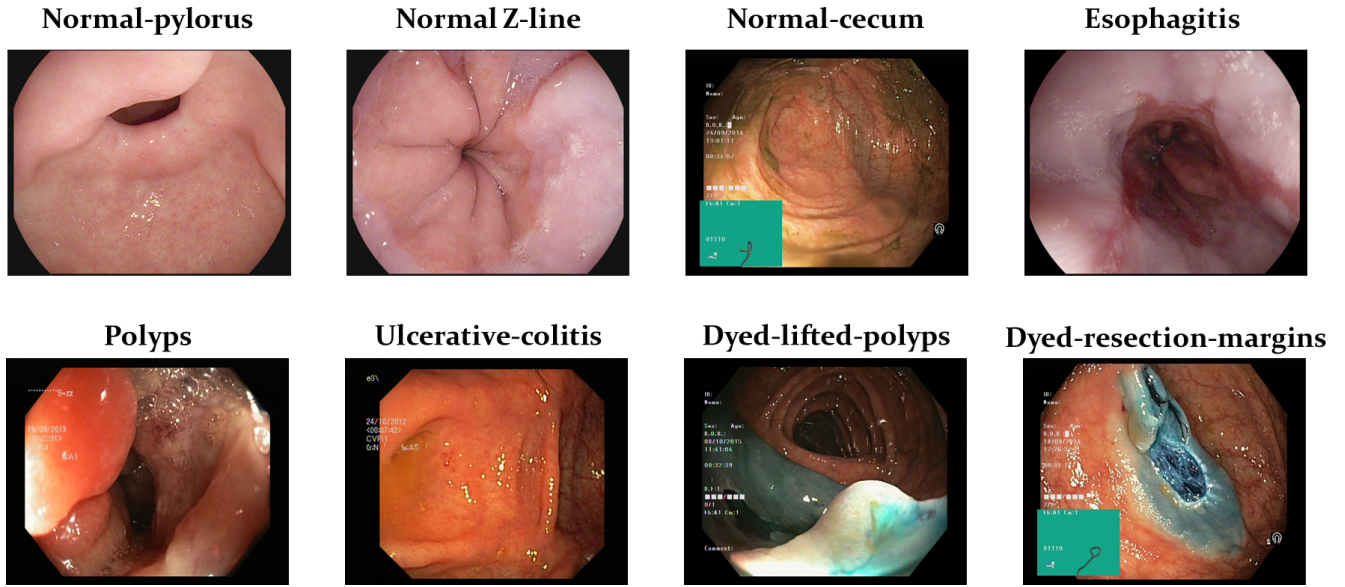

Fig. 1 Representative Images from Each Class in the Kvasir Dataset.

### 4.2 Training Configuration

All models were initialized with weights pre-trained on the ImageNet dataset to leverage transferable low-level visual representations. The ResNet-50 backbone, with its residual connections and deep hierarchical structure, was used as a convolutional baseline known for its strong local feature extraction capabilities. In contrast, the Swin Transformer served as a transformer-based baseline, utilizing hierarchical self-attention over shifted windows to effectively capture both fine-grained textures and long-range dependencies. Additionally, a hybrid architecture combining both ResNet and Swin Transformer was included in the comparison. In this model, features extracted independently from the ResNet and Swin branches were concatenated to form a unified representation. This combined feature vector was then passed through a projection head consisting of a linear layer (input dimension: 2×num_classes, output: 256), followed by ReLU activation, dropout ($p = 0.5$), and a final fully connected layer for class prediction.

### 4.3 Loss Function and Class Imbalance Handling

To address the class imbalance inherent in the Kvasir dataset, we adopted the Class-Balanced Loss formulation, which reweights each class based on the effective number of samples rather than raw class frequencies. Specifically, the class weight is computed as $(1 - \beta)/(1 - \beta^{n_c})$, where $n_c$ is the number of samples in class $c$ and $\beta$ is a smoothing hyperparameter. We set $\beta = 0.999$ to softly down-weight the majority classes while preserving overall loss stability. This approach, originally proposed by Cui et al. [19], mitigates the bias toward majority classes and enhances the model's sensitivity to minority categories, contributing to a more equitable and stable learning process. In our experiments, it proved effective in improving recognition of underrepresented classes such as pylorus and retroflexed stomach in the Kvasir dataset.

### 4.4 Optimization Strategy and Training Protocol

All models were trained using the AdamW optimizer, with an initial learning rate of 0.001, batch size of 16, and for a total of 20 epochs. To enhance generalization and reduce overfitting, dropout and weight decay were employed as regularization techniques. Additionally, a comprehensive data augmentation pipeline was applied during training to simulate the variability commonly observed in clinical imaging conditions. Model checkpoints were selected based on the highest Matthews Correlation Coefficient (MCC) achieved on the validation set, ensuring a robust evaluation criterion particularly suited for imbalanced multi-class settings. All experiments were conducted using a single NVIDIA RTX 3080 Ti GPU.

### 4.5 Evaluation Metrics

Model performance was assessed using a comprehensive suite of classification metrics, including accuracy, macro- averaged precision, recall, F1-score, and the Matthews Correlation Coefficient (MCC). Given the multi-class nature of the classification task and the presence of class imbalance, these metrics were selected to provide a holistic view of the model's behavior across all classes. In particular, macro averaging was employed to compute each metric independently for each class before taking the unweighted mean, ensuring equal treatment of both frequent and infrequent classes. Micro averaging, in contrast, was also considered to reflect overall model performance by aggregating contributions from all classes, thus favoring more prevalent labels. Among these, MCC was selected as the primary evaluation metric due to its robustness in imbalanced multi-class classification scenarios and its ability to account for all elements of the confusion matrix. Furthermore, confusion matrices were analyzed to assess class-specific prediction quality, revealing dominant misclassification patterns and identifying performance bottlenecks that guided the interpretation of empirical results.

### 4.6 Open Set Recognition Hyperparameter Optimization

To enhance open set recognition performance, we conducted hyperparameter optimization for both the Softmax Thresholding and OpenMax methods using the Optuna framework. The optimization was carried out on the validation set, with classification accuracy as the objective to maximize. For Softmax Thresholding, we optimized the decision threshold in the range of 0.50 to 0.95, aiming to reject unknown classes more effectively while maintaining classification performance on known classes. For OpenMax, a more extensive set of hyperparameters was tuned: the Weibull tail size was explored in the range 20 to 400, the alpha parameter (number of top classes used for calibration) from 1 to min(10, number of known classes), and the rejection threshold from 0.60 to 0.99. These hyperparameters critically affect the fitting of the Weibull distribution and the decision boundary between known and unknown categories. The results of applying the optimized methods are summarized in Table 3, which compares the performance of ResNet-50, Swin Transformer, and HybridSwinResNet on the Kvasir dataset using Open Set Recognition (OSR) with Softmax, Softmax Thresholding, and OpenMax approaches. Among the models, ResNet-50 with OpenMax achieved the highest overall performance, recording 86.3% accuracy, 94.7% AUROC, and 70.4% AUPR-OUT. These results demonstrate its strong capability to distinguish unknown samples, supported by balanced precision and recall values (85.4% / 85.5% macro/micro precision). The Hybrid SwinResNet model also showed consistently strong performance, especially with OpenMax, achieving 83.5% accuracy, 89.4% AUROC, and 56.4% AUPR-OUT. Notably, Hybrid SwinResNet exhibited significant improvements across all metrics compared to its vanilla Softmax baseline (75.0% accuracy, 97.2% AUROC, and 51.5% AUPR-OUT), indicating that architectural fusion contributes to robustness in OSR tasks. In contrast, the Swin Transformer underperformed across all OSR methods. With OpenMax, its accuracy reached only 63.7%, and AUPR-OUT remained low at 25.8%, suggesting limited ability to reject unknown classes effectively. This may reflect sensitivity to overfitting or insufficient calibration in the transformer-based feature representations. Overall, these findings underscore the effectiveness of OpenMax over Softmax-based methods when handling open set scenarios, particularly when combined with suitable architectures and hyperparameter tuning. The improvements observed in AUPR-OUT highlight OpenMax's superior discrimination of unknown classes, which is crucial for safe and reliable deployment in medical image classification tasks such as those using the Kvasir dataset.

Table 1 Hyperparameter Search Ranges by Method

| Method | Hyperparameter | Search Range |
|---|---|---|
| Softmax Threshold | Softmax_threshold | 0.5 ~ 0.95 |
| OpenMax | Weibull_tail | 20 ~ 400 |
|  | Weibull_alpha | 1 ~ 3 |
|  | Weibull_threshold | 0.6 ~ 0.99 |

## 5. Results and discussion

To establish a performance baseline prior to open set evaluation, we first assessed the closed-set classification perfor- mance of three different models: ResNet-50, Swin Transformer, and our proposed HybridSwinResNet. All models were trained and tested on the Kvasir dataset under the assumption that all test samples belong to one of the known training classes. The HybridSwinResNet model achieved the best overall performance among the three, with 99.7% accuracy, 99.7% precision (macro/micro), 99.7% recall (macro/micro), 99.7% F1-score (macro/micro), and a Matthews Correlation Coefficient (MCC) of 99.5%. This indicates that the hybrid architecture, which combines feature repre- sentations from both ResNet and Swin Transformer, effectively leverages the strengths of each backbone to produce highly discriminative features for endoscopic image classification. The ResNet-50 model also demonstrated excellent results, with 99.3% accuracy, 99.0% precision, 99.0% recall, 99.0% F1-score, and an MCC of 99.0%, confirming the reliability and robustness of ResNet-based convolutional architectures in medical imaging tasks. In contrast, the Swin Transformer model yielded relatively lower performance, with 89.7% accuracy, 89.6% macro / 89.0% micro precision, 89.0% recall, 89.0% F1-score, and an MCC of 84.6%. While these values are lower than those of the other models, the Swin Transformer still provides reasonably good classification performance and suggests potential for improvement through fine-tuning or architectural enhancements. Overall, the results demonstrate that HybridSwinResNet not only surpasses the individual backbones in all key evaluation metrics but also maintains a high level of balance across performance categories. This balanced performance indicates that the hybrid model may be particularly well-suited for generalization to unseen classes, a critical factor for open set recognition tasks.

Table 2 Performance Metrics of SOTA Models on Closed Set Classification Using on the Kvasir Dataset

| Model | Metrics (%, Macro/Micro) | | | | |
| --- | --- | --- | --- | --- | --- |
| | Acc. | Precision | Recall | F1-Score | MCC |
| ResNet-50 | 99.3 | 99.0/99.0 | 99.0/99.0 | 99.0/99.0 | 99.0 |
| Swin Transformer | 89.7 | 89.6/89.0 | 89.0/89.0 | 89.0/89.0 | 84.6 |
| Hybrid SwinResNet | 99.7 | 99.7/99.7 | 99.7/99.7 | 99.7/99.7 | 99.5 |

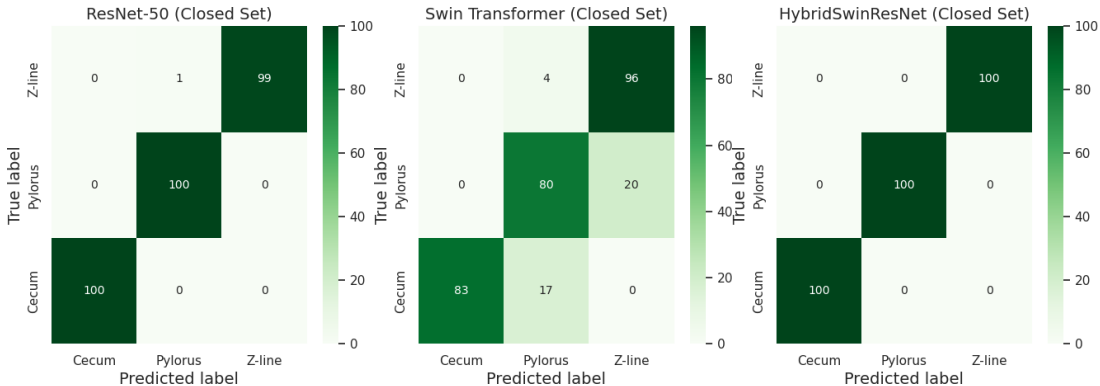

Fig. 2 Closed-set Confusion Matrices on the Kvasir Dataset.

The confusion matrices revealed that some misclassifications occurred between visually similar classes such as esophagitis and normal-z-line, which aligns with the known difficulty of distinguishing between subtle mucosal changes in gastrointestinal imaging. Despite this, most classes—especially anatomically distinct regions like pylorus and ulcerative colitis—were classified with high confidence. Macro-averaged classification reports showed that all models achieved high precision, recall, and F1-scores, generally exceeding 0.90 for most categories. These results affirm the effectiveness of deep convolutional and transformer-based models in closed-set medical image classification. More importantly, they provide a reliable benchmark for evaluating performance under open set conditions in the next stage of our analysis. For the open-set evaluation, we followed a realistic clinical setup by defining a closed-set training scenario with three known classes: normal-cecum, normal-pylorus, and normal-z-line. The remaining five classes—dyed-lifted- polyps, dyed-resection-margins, esophagitis, polyps, and ulcerative-colitis—were excluded from training and treated as unknown during testing. This division allowed us to evaluate each model's ability to both correctly classify known samples and reliably detect or reject novel, unseen classes, which is critical in open-set recognition scenarios involving medical image data.

Table 3 Performance Metrics of SOTA Models on Open Set Recognition Using OpenMax on the Kvasir Dataset

| Model | Method | Metrics (%, Macro/Micro) | | | | | |
|---|---|---|---|---|---|---|---|
| | | Acc. | Precision | Recall | F1-Score | AUROC | AUPR-OUT |
| ResNet-50 | Softmax | 74.5 | 57.4/74.5 | 74.5/74.5 | 64.4/74.5 | 98.6 | 67.0 |
| | Softmax Threshold | 82.3 | 84.3/81.0 | 81.0/81.0 | 77.6/81.0 | 98.6 | 67.0 |
| | OpenMax | 86.3 | 85.4/85.5 | 85.5/85.5 | 85.3/85.5 | 94.7 | 70.4 |
| Swin Transformer | Softmax | 64.2 | 48.7/64.2 | 64.3/64.2 | 55.2/64.2 | 87.0 | 26.3 |
| | Softmax Threshold | 63.2 | 54.1/63.2 | 63.2/63.2 | 56.2/63.2 | 87.0 | 26.3 |
| | OpenMax | 63.7 | 63.1/63.7 | 63.8/63.7 | 63.2/63.7 | 81.4 | 25.8 |
| Hybrid SwinResNet | Softmax | 75.0 | 57.9/75.0 | 75.0/75.0 | 64.9/75.0 | 97.2 | 51.5 |
| | Softmax Threshold | 78.5 | 83.7/78.5 | 78.5/78.5 | 72.7/78.5 | 97.2 | 51.5 |
| | OpenMax | 83.5 | 84.6/83.5 | 83.5/83.5 | 81.9/83.5 | 89.4 | 56.4 |

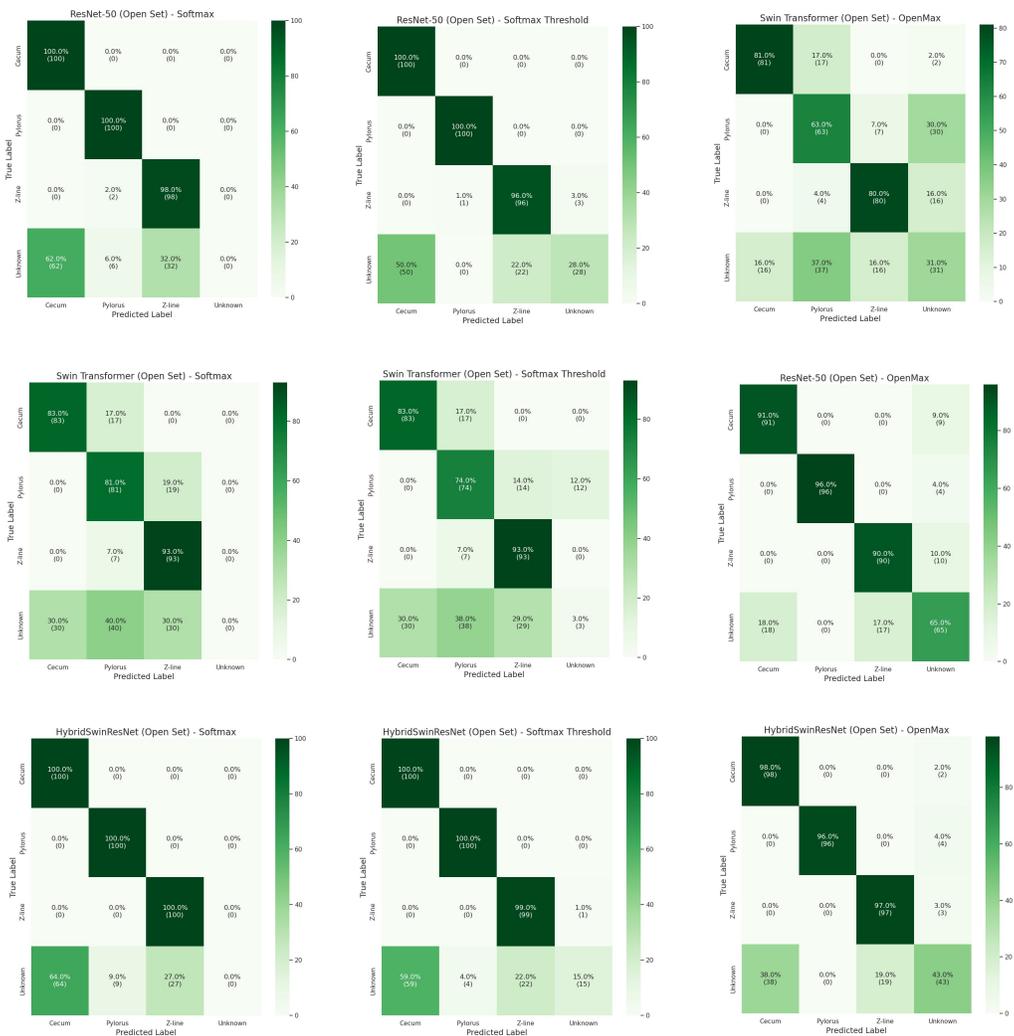

Fig. 3 Comparison of Confusion Matrices Across OSR Methods for All Models on the Kvasir Dataset

# 6. Conclusion

In this study, we investigated the application of Open Set Recognition (OSR) techniques for endoscopic image classification using the Kvasir dataset. While conventional deep learning models such as ResNet-50 and Swin Transformer have shown strong performance in closed-set scenarios, their reliability diminishes when exposed to previously unseen classes—a common occurrence in real-world clinical settings. To address this challenge, we implemented and evaluated both Softmax Thresholding and OpenMax approaches across three different architectures:

ResNet-50, Swin Transformer, and a proposed HybridSwinResNet model that combines convolutional and transformer- based representations. Our experimental results demonstrate that OpenMax, when carefully optimized, significantly improves the model's ability to reject unknown classes while maintaining classification performance on known categories. Notably, ResNet-50 with OpenMax achieved the highest overall OSR performance, while the HybridSwinResNet model offered a balanced trade-off between generalization and discrimination. In contrast, the Swin Transformer alone underperformed in open-set conditions, indicating the need for further tuning or architectural enhancements for robust OSR. This work represents one of the first systematic efforts to explore OSR in gastrointestinal endoscopy using the Kvasir dataset, providing a valuable benchmark for future research. Our findings highlight the critical importance of OSR techniques in ensuring the safe and reliable deployment of AI systems in medical imaging applications, where the presence of unknown or novel conditions is inevitable. By bridging the gap between closed-set training assumptions and open-world clinical realities, this study contributes toward the development of more resilient and trustworthy diagnostic tools for endoscopic examination.